\documentclass{article}

\usepackage{arxiv}

\usepackage[utf8]{inputenc} % allow utf-8 input
\usepackage[T1]{fontenc}    % use 8-bit T1 fonts
\usepackage{hyperref}       % hyperlinks
\usepackage{url}            % simple URL typesetting
\usepackage{booktabs}       % professional-quality tables
\usepackage{amsfonts}       % blackboard math symbols
\usepackage{nicefrac}       % compact symbols for 1/2, etc.
\usepackage{microtype}      % microtypography
\usepackage{lipsum}		% Can be removed after putting your text content
\usepackage{graphicx}
\usepackage{natbib}
\usepackage{doi}

\title{An Analysis on Large Language Models in Healthcare: A Case Study of BioBERT}

\author{Shyni Sharaf\\
	School of Computer Science and Engineering\\
	Kerala University of Digital Sciences-\\Innovation and Technology\\
	Thiruvananthapuram, India \\
	\texttt{shyni.cs22@duk.ac.in} \\
        \And
	V. S. Anoop \\
        Applied NLP Research Lab\\
	School of Digital Sciences\\
	Kerala University of Digital Sciences-\\Innovation and Technology\\
	Thiruvananthapuram, India \\
	\texttt{anoop.vs@duk.ac.in} \\
}

\date{}

\hypersetup{
}

\begin{document}
\maketitle

\begin{abstract}
This paper conducts a comprehensive investigation into applying large language models, particularly on BioBERT, in healthcare. It begins with thoroughly examining previous natural language processing (NLP) approaches in healthcare, shedding light on the limitations and challenges these methods face. Following that, this research explores the path that led to the incorporation of BioBERT into healthcare applications, highlighting its suitability for addressing the specific requirements of tasks related to biomedical text mining. The analysis outlines a systematic methodology for fine-tuning BioBERT to meet the unique needs of the healthcare domain. This approach includes various components, including the gathering of data from a wide range of healthcare sources, data annotation for tasks like identifying medical entities and categorizing them, and the application of specialized preprocessing techniques tailored to handle the complexities found in biomedical texts. Additionally, the paper covers aspects related to model evaluation, with a focus on healthcare benchmarks and functions like processing of natural language in biomedical, question-answering, clinical document classification, and medical entity recognition. It explores techniques to improve the model's interpretability and validates its performance compared to existing healthcare-focused language models. The paper thoroughly examines ethical considerations, particularly patient privacy and data security. It highlights the benefits of incorporating BioBERT into healthcare contexts, including enhanced clinical decision support and more efficient information retrieval. Nevertheless, it acknowledges the impediments and complexities of this integration, encompassing concerns regarding data privacy, integrity, bias mitigation, transparency, resource-intensive requirements, and the necessity for model customization to align with diverse healthcare domains.
\end{abstract}

% keywords can be removed

\section{Introduction}
NLP processing has evolved to become LLM (Large Language Model). In the 1940s, after World War II, people realized the importance of translation between languages and wanted to create a machine that could perform automatic translation. Early NLP systems were rule-based systems that humans manually programmed with rules for processing language. These systems often had many limitations in handling complex language and could be easily deceived by unexpected inputs. In the early 2000s, statistical NLP models began to emerge. These models were trained on large text datasets and learned to predict the next word in a sequence based on preceding words. Statistical NLP models exhibited greater robustness compared to rule-based systems\cite{wang2023prompt} and could handle a wider range of language tasks. By the mid-2010s, deep learning models started to revolutionize NLP. These models, based on artificial neural networks, had the capability to learn intricate patterns from data. Deep learning models \cite{lavanya2021deep}quickly outperformed statistical NLP models such as machine translation, summarization of texts, and question answering. In 2017, the Transformer \cite{la2022transformer}a deep learning model designed for processing sequential data, such as text, marked a significant milestone. It achieved state-of-the-art results across numerous NLP tasks and soon became the standard architecture for training LLMs. This shift led to the emergence of natural language processing (NLP) as we know it today. NLP has radically transformed into the era of large language models (LLMs).
\par LLMs are trained on massive text datasets, sometimes containing hundreds of billions or even trillions of words. This extensive training enables LLMs to understand the intricate patterns and relationships inherent in language. As a result, LLMs have fundamentally altered how we interact with and harness the power of language. Popular LLMs like GPT-3.5, GPT-4, PaLM, Cohere, LaMDA, Llama, and others have revolutionized our interaction with data by redefining the boundaries of language understanding and generation. Natural Language Processing (NLP) is a branch of artificial intelligence (AI) and computational linguistics that facilitates interaction between computers and humans through natural language. LLMs\cite {reddy2023evaluating}process vast amounts of textual data, learn the underlying patterns, and generate contextually relevant human-like text. This technology has not only catalyzed but has become a driving force in transforming healthcare and biomedical applications.
\indent In this article, we conduct a comparative analysis of the diverse applications of LLMs in the healthcare and biomedical domains. We explore how LLMs are reshaping the landscape by offering innovative solutions to long-standing challenges. Current healthcare and biomedical systems often operate inefficiently, have limited access to relevant information, and involve cumbersome documentation processes. LLMs can address these challenges by providing rapid, context-aware responses to medical queries, extracting valuable insights from unstructured data, and automating clinical documentation. The major contributions of this article are as follows:
\begin{itemize}
    \item Conducts a detailed evaluation of the existing prominent state-of-the-art large language models introduced in the healthcare domain.
    \item Taking BioBERT as a reference pre-trained language model, we check the applications of the same in healthcare.
    \item Discusses the prominence of BioBERT in downstream clinical natural language processing tasks and discusses them in detail.
    \item Outlines the challenges with LLM in healthcare and presents the future research directions.
\end{itemize}

\subsection{An overview of LLM components}
\begin{itemize}
    \item \textbf{Input Text}: Initially, the LLM receives raw text as input. This text can be in sentences, paragraphs, or documents.
    \item \textbf{Tokenization}: The text we give is divided into individual tokens. Tokens can be words, subwords, or characters, depending on LLMs tokenization scheme. This step breaks down the text into manageable units for processing.
    \item \textbf{Word Embeddings}: Each token is transformed into a high-dimensional vector through word embeddings. These vectors capture the token's meaning and context. Word embeddings are learned during the model's training using a vast amount of text data.
    \item \textbf{Transformer Layers}: The embedded vectors are passed through multiple transformer layers. Each transformer layer consists of two main components:
        \begin{itemize}
            \item Multi-Head Self-Attention: This component weighs the importance of each token in relation to others, capturing dependencies and context.
            \item Feedforward Neural Networks: Complex transformations are applied to the vectors, enhancing the model's understanding of patterns and relationships.
        \end{itemize}
    \item \textbf{Output Layer}: After processing through the transformer layers, the output is fed into a linear layer. This linear layer generates a probability distribution that spans the model's vocabulary. It estimates the probability of different words or word sequences following the input text.
    \item \textbf{Probability Estimation}: The probability distribution generated in the previous step is used for various tasks such as language generation, text completion, and question answering.
    \item \textbf{Training and Fine-Tuning}: LLMs are initially trained on a large text corpus to learn embeddings and model parameters. Fine-tuning can follow, where the model is further trained on task-specific data to adapt its language understanding to the specific task or domain.
\end{itemize}

\section{NLP for Healthcare Applications}
LLMs have emerged as a transformative technology in healthcare, enabling an extensive range of applications, from clinical decision support to medical data analysis. LLMs allow healthcare professionals to harness the power of language data for improved patient care, research, and administrative tasks.
\begin{itemize}
    \item \textbf{Medical Question Answering}: LLMs can answer medical questions, providing quick and accurate responses. This application aids healthcare professionals in accessing medical knowledge and information rapidly.
    \item \textbf{Electronic Health Record (EHR) Analysis}: LLMs can analyze unstructured text in electronic health records, extracting valuable insights about patient histories, diagnoses, treatments, and clinical notes. This supports clinical decision-making and research.
    \item \textbf{Clinical Documentation}: LLMs can assist healthcare providers in generating clinical notes, reports, and documentation. This streamlines the documentation process, allowing clinicians to focus more on patient care.
    \item \textbf{Medical Imaging}: LLMs can assist in medical image interpretation by generating natural language descriptions of images. This can improve communication between radiologists and referring physicians\cite{wang2023chatcad}\cite{rao2023evaluating}.
    \item \textbf{Clinical Decision Support}: LLMs can provide context-aware information to support clinical decisions. They can recommend treatment options, predict patient outcomes, and identify potential risks \cite{rao2023evaluating}.
     \item \textbf{Healthcare Communication}: LLMs can improve doctor-patient communication by offering language translation services, ensuring effective communication in multilingual healthcare settings \cite{yunxiang2023chatdoctor}
     \item \textbf{Patient Engagement}: LLMs can be used in chatbots and virtual assistants \cite{ray2023chatgpt} to engage with patients, answer their healthcare queries, and provide health-related information and guidance. Healthcare professionals can use NLP to extract relevant information from patient records, such as medical history, medication allergies, and previous diagnoses, enabling the creation of personalized treatment plans and early identification of high-risk patients for disease prevention.
     \item \textbf{Enhancing Medical Research}: NLP can also analyze large amounts of medical data to identify patterns and trends\cite{hao2018bibliometric}. It helps researchers to develop new treatments and therapies.
     \item \textbf{Improving Clinical Trials}: NLP algorithms can sift through much data and extract information relevant to the clinical trial. NLP helps clinical trials \cite{chen2020trends}by finding the right participants faster and cheaper through patient data analysis and improves efficiency. It reduces the time and cost.
     \item \textbf{Improving Digital Health Records}: NLP can make digital patient records more correct and complete. These records hold information about a person’s health history and treatments. NLP helps doctors to get the right details from these health records \cite{costea2020machine}so they can make better decisions for patient care.
     \item \textbf{Supports Medical Practitioners}: NLP makes many everyday tasks of health professionals easier\cite{demner2009can}. For instance, it finds possible issues with medicines, helps doctors adjust treatment plans, and helps doctors write notes faster by saving time and reducing errors, so they can spend more time caring for patients.
 Also, NLP aids in extracting Information from medical literature, helping healthcare professionals to learn \cite{henwood2007nlp}stay current with the latest research and best practices.
\end{itemize}

\section{Language Models and Healthcare}
Large Language Models (LLMs) are one of the most exciting areas in Artificial Intelligence (AI) research that have the ability to process and generate human-like text for various healthcare applications. The more data we train, the more predictions will be more accurate. Mainly used LLM are GPT-3, BERT, and RoBERTa \cite{liu2019roberta}, which are trained on billions of words and patterns. so these models can understand the structure easily and generate text. Once the model is generated it can be fine-tuned for a specific Task. The applications of LLM \cite{hao2018bibliometric} healthcare have many different aspects and have the power to bring about significant positive changes in various fields. These technologies offer real-time assistance to healthcare professionals by helping them diagnose diseases and give the right treatments without Errors. Predictive Analytics in healthcare can use data to predict disease outbreaks and enhance healthcare delivery efficiency. The significance of studying LLM applications in healthcare is that it is versatile. LLMs should be used in healthcare in a collaborative and verified way to ensure responsible and effective use, ultimately improving patient care. This means the use of LLMs should be carefully monitored and evaluated, and thereby identifying potential problems or risks.
LLMs are a powerful tool that has the potential to revolutionize healthcare.

\subsection{Benefits of LLMs in Healthcare:}
\begin{itemize}
    \item \textbf{Improved Support for Clinical Decisions}: LLMs assist healthcare providers in decision-making by providing access to a vast amount of medical knowledge and up-to-date research. They can suggest potential diagnoses, treatment options, or relevant research articles quickly. LLMs can make the diagnosis and data more accurate than humans; thereby, the quality of outcomes and care of patients can be improved.
    \item \textbf{Efficient Information Retrieval}: LLMs can be used to clarification medical tests by analyzing the results providing valuable information and helping to find out the abnormalities. By this, we can reduce the time and cost of interpreting the results and improve accuracy and reliability.
    \item \textbf{Clarification of Medical Tests}: LLMs can identify clinical trials by analyzing the current conditions of the patient, medical history, and treatment plans. This will improve the efficiency and effectiveness and provide potential lifesaving treatments.
    \item \textbf{Searching for Potential Clinical Trials}: LLMs can identify clinical trials by analyzing the current conditions of the patient, medical history, and treatment plans. This will improve the efficiency and effectiveness and provide potential lifesaving treatments
\end{itemize}

\subsection{Limitations and Challenges of Using LLMs in Healthcare}
\begin{itemize}
    \item \textbf{Data Privacy and Security}: Integrating LLMs into healthcare must proceed cautiously to safeguard highly sensitive healthcare data. Ensuring data privacy and security, along with compliance with regulations such as HIPAA, is paramount to prevent the potentially severe consequences of data breaches.
    \item \textbf{Bias and Fairness}: LLMs trained on biased data may produce biased or unfair results in healthcare applications. This can lead to disparities in care, misdiagnoses, or unfair allocation of resources.
    \item \textbf{Lack of Transparency}: LLMs often operate as "black boxes," making it challenging to understand their decision-making processes. This lack of transparency can hinder trust among healthcare professionals and patients.
    \item \textbf{Quality Control}: Ensuring the quality and accuracy of information generated or retrieved by LLMs is crucial. Erroneous information or recommendations could harm patients or mislead healthcare providers.
    \item \textbf{Concern of Ethical issues}: Using LLMs in healthcare raises ethical concerns, such as the potential for technology to replace human interaction in patient care, leading to depersonalized medicine.
    \item \textbf{Resource Intensiveness}: Developing, fine-tuning, and maintaining LLMs for healthcare can be resource-intensive regarding computational power, data annotation, and expert oversight.
    \item \textbf{Generalization Challenges}: LLMs may struggle with generalizing to specific healthcare domains, specialties, or rare conditions if not adequately fine-tuned. Customization may be necessary.
\end{itemize}

\section{Related Studies}
In this section, we review relevant articles that have explored the integration of LLMs in healthcare applications. The articles showcase the significant impact of large language models in various healthcare-related tasks, such as biomedical text mining, medical image interpretation, medical question answering, and processing electronic health records. They also highlight the need for careful evaluation and consideration of limitations when applying these models in clinical settings. The strengths include improved task performance and potential benefits for healthcare However, the resource-intensive nature of such models and potential challenges in fine-tuning for specific healthcare applications should be considered. N. Kang et al. \cite{kang2013using}primarily focus on evaluating the performance of MetaMap and Peregrine tools used for biomedical concept normalization. The study investigates the usefulness of rule-based NLP modules that are used to enhance the performance of MetaMap and Peregrine, an adjunct to dictionary-based concept normalization in the biomedical field, to evaluate the Corpus for Arizona Disease.\\
\indent S. A. Hasan et al. \cite{hasan2019clinical} discuss the application of deep learning(DL) techniques in clinical natural language processing (CNLP). The model emphasizes the use of DL models for various clinical applications. Deep learning-driven clinical NLP applications include diagnostic inferencing, biomedical article retrieval, clinical paraphrase generation, adverse drug event detection, and medical image caption generation. J. Lee et al. lee2020biobert\cite{lee2020biobert}introduced BioBERT, a pre-trained biomedical language representation model tailored for biomedical text mining. BioBERT's training involved a substantial biomedical text corpus. This model excelled in various tasks such as named entity recognition, relation extraction, and question answering, achieving state-of-the-art performance across the board.\\
\indent Shin et al.\cite{shin2020biomegatron}contributed to the field with BioMegatron, a larger pre-trained biomedical language model aimed at biomedical text mining analogous to BioBERT. Differing in scale, BioMegatron was trained on an even more extensive corpus of biomedical text and exhibited state-of-the-art performance in tasks such as entity recognition, relation extraction, and question-answering. Additionally, X. Yang et al.\cite{yang2022large} presented GatorTron, a substantial clinical language model created for processing and interpreting electronic health records (EHRs). With extensive scaling in model parameters and training data, GatorTron significantly improved performance across clinical NLP tasks, offering potential enhancements in healthcare-related natural language processing by evaluating it on 5 clinical NLP tasks like clinical concept extraction, medical relation extraction, semantic textual similarity, natural language inference (NLI), and medical question answering (MQA).\\
\indent J.Singhal et al.\cite{singhal2022large}explored the encoding of clinical knowledge using Large Language Models. They demonstrated that training LLMs on extensive clinical text enabled them to accurately answer questions related to clinical concepts, showcasing their potential in encoding clinical knowledge. They proposed a robust framework for human evaluation of model responses, incorporating factors such as factuality, precision, potential harm, and bias into the assessment process. PaLM, and its instruction-tuned variant, Flan-PaLM, were evaluated using MultiMedQA. Wang et al.\cite{wang2023chatcad}presented Chatcad, a large language model designed for interactive computer-aided diagnosis (CAD) in medical image analysis. Trained on a dataset featuring medical images and their accompanying text descriptions, Chatcad demonstrated the ability to accurately diagnose diseases from images, aiding radiologists in their diagnoses.\\ 
\indent S. Reddy et al.\cite{kang2013using}reddy2023evaluatingintroduced a framework for evaluating the translational value of Large Language Models (LLMs) in healthcare. This framework was a comprehensive tool for assessing LLMs' performance in healthcare applications. It was subsequently employed to assess the NLP performance of LLM’s on the grounds of not assessing the models' functional, utility, and ethical aspects as they apply to healthcare, and recommended governance aspects of LLMs in healthcare are required. Zhang et al.\cite{zhang2023huatuogpt}unveiled HuatuoGPT, a specialized LLM tailored for medical consultation. By leveraging data from ChatGPT and real-world doctors, HuatuoGPT was fine-tuned to provide clinical advice and support to patients. This unique approach improved its performance, achieving state-of-the-art results in medical consultation tasks. K. Singhal et al.\cite{singhal2023towards}introduced Med-PaLM 2, an LLM designed for expert-level medical question answering. This model achieved remarkable scores in medical question-answering tasks with a score of 67.2\% on the MedQA dataset, highlighting its potential for delivering high-precision performance to medical question answering.
\begin{table}[]
\caption{Some state-of-the-art approaches in using LLM and related techniques in the healthcare NLP}
\label{tab:my-table}
\begin{tabular}{@{}llp{3in}l@{}}
\toprule
Author             & Model                                                                            & Methodology                                                                                                                                                                                                        \\ \midrule
N. Kang et al. \cite{kang2013using}     & Rule-based NLP                                                                   & A rule-based NLP module used to enhance the performance of MetaMap and Peregrine.                                                                                                                                  \\
S. A. Hasan et al. \cite{hasan2019clinical} & Deep Learning                                                                    & Addresses the challenges posed by clinical documents, including acronyms, nonstandard clinical jargon, inconsistent document structure, and privacy concerns.                                           \\
J. Lee et al. \cite{lee2020biobert}  & BioBERT                                                                          & Pre-training on large-scale biomedical corpora outperforms BERT and other models in biomedical text-mining tasks.                                                                                                 \\
HC Shin et al. \cite{shin2020biomegatron}     & BioMegatron                                                                      & Empirical study on factors affecting domain-specific language models, pre-training on larger domain corpus.                                                                                                        \\
X. Yang et al. \cite{yang2022gatortron}     & GatorTron                                                                        & Developing a large clinical language model, scaling up the number of parameters and training data.                                                                                                                 \\
K. Singhal et al.\cite{singhal2022large} & \begin{tabular}[c]{@{}l@{}}MultiMedQA, PaLM, \\ Flan-PaLM, Med-PaLM\end{tabular} & MultiMedQA benchmark, human evaluation of model answers, instruction prompt tuning.                                                                                                                                \\
Wang et al.\cite{wang2023chatcad} [5]        & ChatGPT, CAD networks                                                            & Integrating LLMs with CAD networks, enhancing output with natural language text.                                                                                                                                   \\
S. Reddy et al.\cite{reddy2023evaluating}    & --                                                                               & Discusses the potential use of Large Language Models (LLMs) in healthcare. Highlights concerns related to misinformation and data falsification. Proposes a framework for evaluation, including human assessments. \\

H. Zhang et al. \cite{zhang2023huatuogpt}    & HuatuoGPT                                                                        & Leveraging distilled data from ChatGPT and real-world data from doctors for medical consultation, reward model training.                                                                                           \\
K. Singhal et al.\cite{singhal2023towards}  & Med-PaLM - 2                                                                     & Improving upon Med-PaLM with base LLM improvements, medical domain fine-tuning, and prompting strategies.                                                                                                          \\ \bottomrule
\end{tabular}
\end{table}

\section{An Analysis of LLMs in Healthcare - A Case Study of BioBERT}
In this section, we delve into the methodologies and outcomes of the aforementioned articles. We assess how LLMs are employed to address healthcare challenges and explore their impact on various aspects of the healthcare industry. The paper "BioBERT: a pre-trained biomedical language representation model for biomedical text mining" by J Lee et al. \cite{lee2020biobert} investigates how the recently introduced pre-trained language model BERT can be adapted for biomedical corpora. In this article, we explore the possibility of Fine-tuning BioBERT for the healthcare domain, which can be a valuable endeavor given its success in biomedical text-mining tasks. To adapt BioBERT for healthcare applications, methodology outlines the steps and considerations for fine-tuning BioBERT for healthcare-specific tasks. It emphasizes the importance of domain expertise, data quality, and ethical considerations in developing robust and reliable healthcare language models. To adapt BioBERT for healthcare applications, the following methodology can be considered:
\begin{itemize}
    \item \textbf{Data Collection}: Gather a comprehensive and diverse dataset from healthcare and biomedical sources. This dataset should include electronic health records (EHRs) \cite{yang2022large}medical literature, clinical notes, medical imaging reports, and other relevant sources. And annotate the data for various healthcare-related tasks, such as medical entity recognition (e.g., disease names, medications, procedures), medical text classification (e.g., diagnosis prediction, disease classification), and medical question-answering\cite{singhal2023towards}.
    \item \textbf{Pre-processing}: Prepare the data by cleaning and formatting it for training. This may involve standardizing medical terms, removing duplicates, removing any errors or inconsistencies in the data, and handling missing values. Then Customize tokenization to accommodate the unique vocabulary and structure of biomedical and clinical texts. Clinical text data often contains specialized vocabulary and structure, so it is important to use a customized tokenizer for this type of data. Hence specialized tokenizers may be needed to handle medical terminology, abbreviations, and symbols. Some common tokenizers for biomedical and clinical text data include:
        \begin{itemize}
          \item The BioBERT tokenizer. This tokenizer is based on the BERT tokenizer but has been customized to handle medical terminology and abbreviations.
          \item The MedTokenizer. This tokenizer is specifically designed for biomedical text data.
          \item The SciBERT tokenizer. This tokenizer is designed for scientific text data, which includes biomedical text.
        \end{itemize}
    
\end{itemize}

\begin{figure}[ht]
    \centering
    \includegraphics[width=0.8\linewidth]{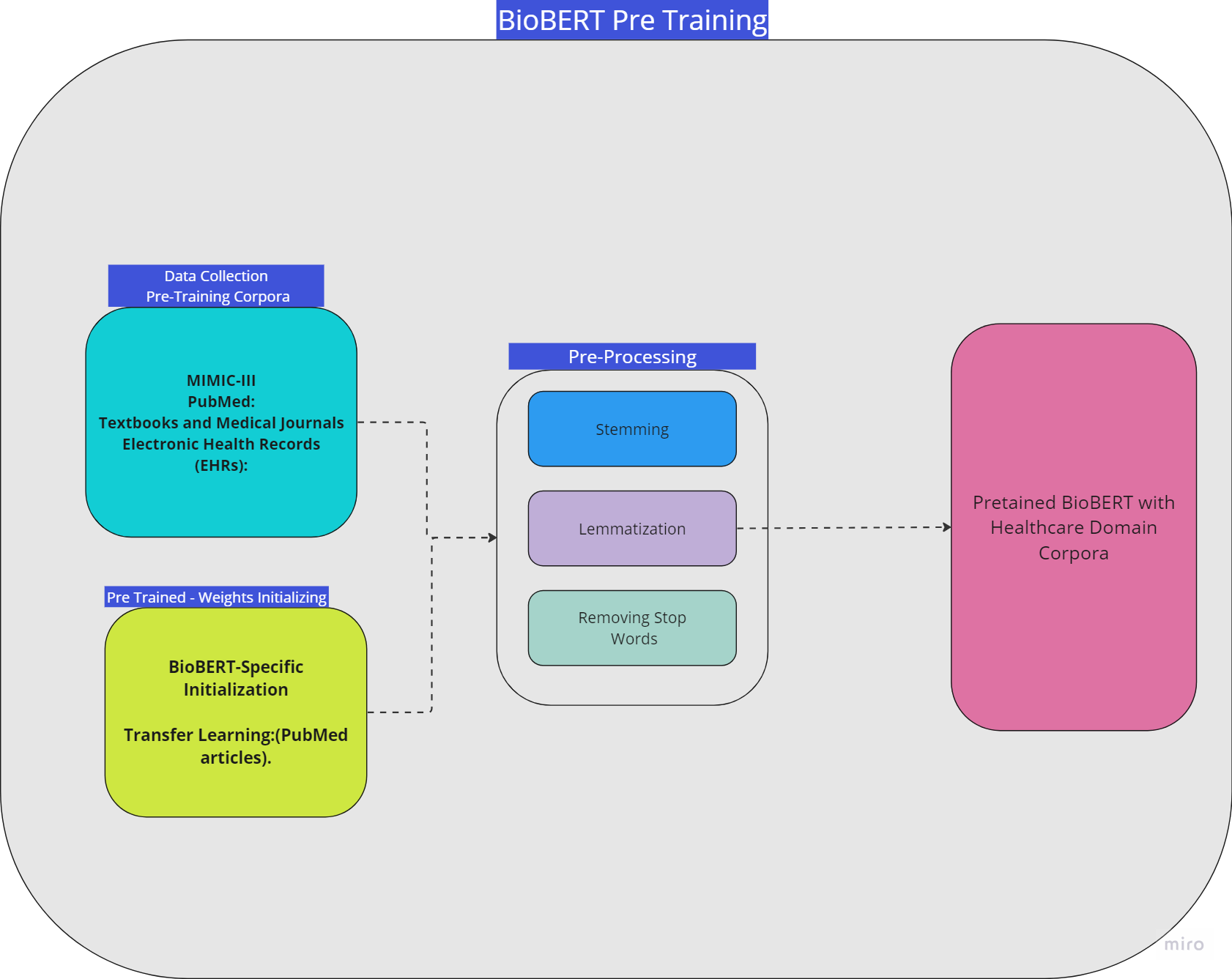} % Replace with the actual image file name
    \caption{Overall architecture for BioBERT Pre-training}
    \label{fig:pretrain}
\end{figure}

\begin{figure}[ht]
    \centering
    \includegraphics[width=0.8\linewidth]{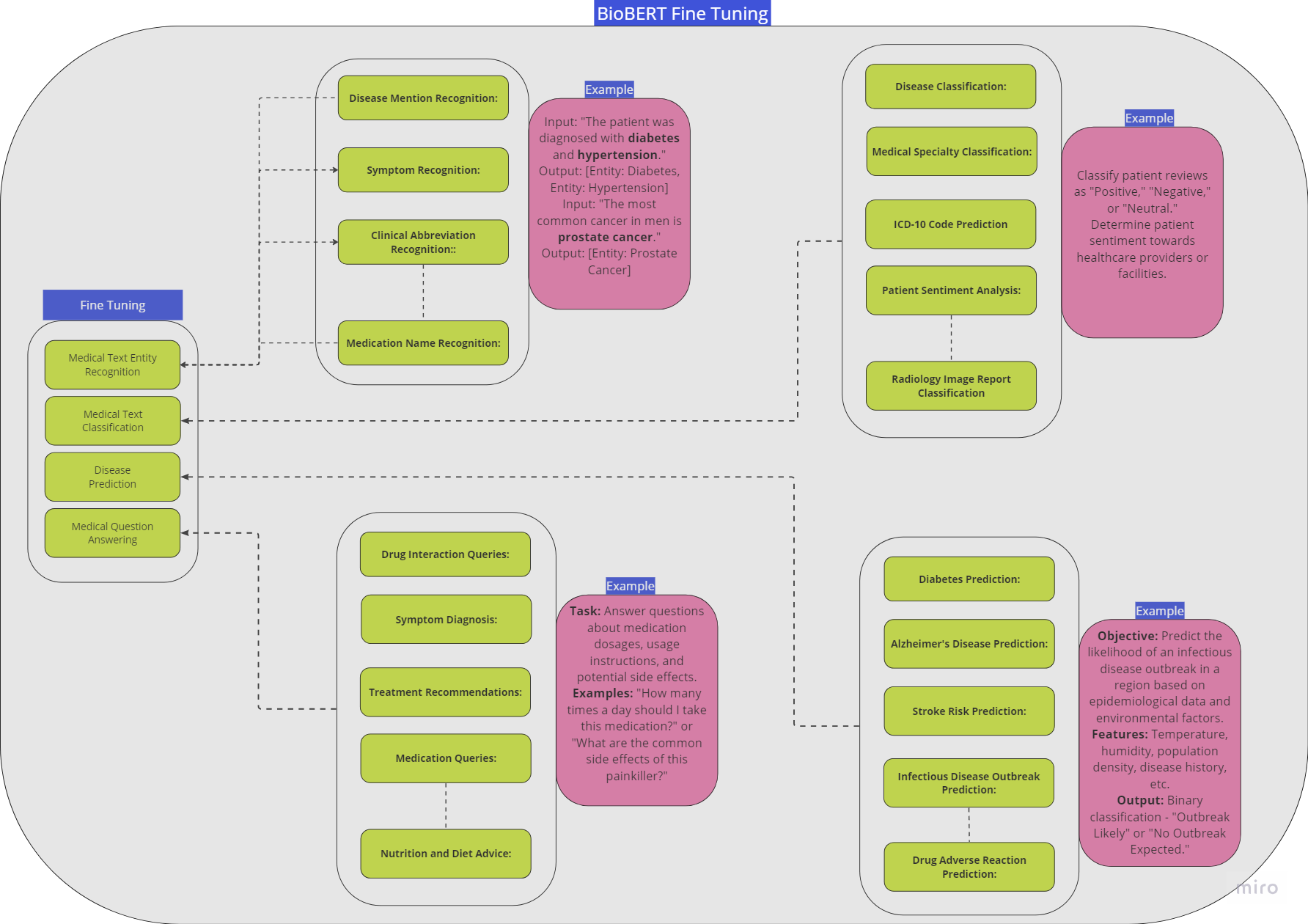} % Replace with the actual image file name
    \caption{Overall architecture for BioBERT Finetuning}
    \label{fig:finetune}
\end{figure}
BioBERT is a pre-trained language model trained on a massive dataset of biomedical text. The pre-trained weights represent the model's understanding of the general structure and semantics of biomedical text. Design a set of downstream tasks specific to healthcare. Many different downstream tasks can be performed using BioBERT. Some common tasks include:
\begin{itemize}
  \item \textbf{Medical Entity Recognition:} This task involves identifying and extracting medical entities from text. These entities can include diseases, medications, and medical procedures.
  
  \item \textbf{Medical Text Classification:} In this task, the text is categorized into different healthcare-related categories. Examples of categories include diagnosis, prognosis, and treatment.
  
  \item \textbf{Disease Prediction:} This task involves predicting the likelihood of a patient having a particular disease.
  
  \item \textbf{Medical Question-Answering:} This task involves answering questions about medical topics based on text.

\end{itemize}
Fine-tune BioBERT on these tasks using the annotated healthcare dataset. Fine-tuning is adjusting the model's weights to improve performance on a specific task. This is done by feeding the model the annotated healthcare dataset and letting it learn from it. Apply appropriate loss functions. A loss function measures the model's performance on the task. The loss function is used to update the model's weights during fine-tuning. Incorporate transfer learning techniques: Transfer learning involves using a model trained on one task to enhance the performance of a model on a distinct task. This can be achieved by initializing the new model with the pre-trained model's weights. Conduct experiments on hyperparameters. Hyperparameters represent the configurations of the machine learning algorithm and significantly influence the model's performance. Common hyperparameters to explore encompass:
\begin{itemize}
 \item Adjust the learning rate, which dictates the magnitude of weight updates in each training iteration.
 \item Vary the batch size, determining the quantity of samples utilized for weight updates in each training iteration.
 \item Modify the number of epochs, specifying how often the model undergoes training on the data.
\end{itemize}

\subsection{Evaluation Metrics:}
Assess the fine-tuned BioBERT model across a range of healthcare-related benchmarks and tasks. These include biomedical NLP tasks, medical question-answering, clinical document classification, medical entity recognition, generating discharge summaries, interpreting medical records, and providing medical advice.

\begin{itemize}
    \item \textbf{F1 score:} It is calculated by taking the harmonic mean of precision and recall. it is the measure of the accuracy and completeness of the model's predictions. The F1 score is a good metric for tasks such as medical entity recognition and text classification.
    \item \textbf{Accuracy:} Accuracy is the percentage of predictions that the model gets correct.
    \item \textbf{Precision:} Precision is the percentage of positive predictions that are actually positive.
    \item \textbf{Recall:} Recall is the percentage of actual positives that are predicted as positive.
    \item \textbf{AUC:} AUC is the area under the receiver operating characteristic curve. It measures the model's ability to distinguish between positive and negative examples. AUC is a good metric for tasks such as medical question-answering and disease prediction.
    \item \textbf{C-index:} The C-index measures the model's ability to predict the survival of patients.
 \end{itemize}

 \subsection{Model Interpretability:}
To enhance the interpretability of a fine-tuned BioBERT model, employ the following techniques:
    \begin{itemize}
     \item \textbf{Analyze the model's predictions : } Examine the model's predictions and comprehend their rationale. This involves inspecting the model's features for making predictions and scrutinizing the attention weights assigned to various parts of the text.
     \item \textbf{Utilize visualization techniques : } Make the model's predictions more comprehensible through graphical representations. Employ heat maps to visualize attention weights or other visualization methods to elucidate how the model generates predictions.
     \item \textbf{Leverage explainability tools : } Utilize various explainability tools designed to elucidate how a machine learning model arrives at its predictions. These tools reveal the features employed by the model for prediction and provide insight into the significance of each feature.
     \end{itemize}
 \subsection{Validation and Testing}
 To validate the performance of a fine-tuned BioBERT model for healthcare tasks, consider the following actions.
    \begin{itemize}
      \item \textbf 	Compare model's performance with that of other existing biomedical models like BioMegatron\cite{shin2020biomegatron} GatorTron \cite{yang2022gatortron}and clinical language models\cite{singhal2022large}. Use the same evaluation metrics and datasets to determine the best-performing model based on these metrics.
       \item \textbf Experiment with hyperparameters, recognizing that these settings can significantly influence the model's performance. Conduct experiments with different hyperparameters to identify the optimal configuration for the specific task.
       \item \textbf Validate the model on external healthcare datasets or benchmarks to assess its generalizability and robustness. The model should demonstrate strong performance on previously unseen datasets.\\
    \end{itemize}
   When validating the performance of a fine-tuned BioBERT model for healthcare tasks, also consider the following factors:
   \begin{itemize}
        \item The size and quality of the training dataset.
        \item The specific task for which the model is being evaluated.
        \item The choice of evaluation metrics.
        \item The clinical requirements that the model aims to address.
    \end{itemize}
\subsection{Deployment and Integration:}
To deploy and integrate a fine-tuned BioBERT \cite{lee2020biobert}model into healthcare applications and systems, take the following actions:
\begin{itemize}
      \item \textbf Apply regularization techniques to prevent overfitting, a potential issue when training the model on a limited dataset. Overfitting occurs when the model captures noise in the data rather than the underlying patterns. Regularization discourages the model from learning non-generalizable patterns.
      \item \textbf 	Augment dataset by artificially increasing its size. Employ techniques such as image translation, text generation, and synthetic data creation to enhance the dataset. Data augmentation bolsters the model's performance by increasing its resilience to data noise and variations.
      \item \textbf 	Integrate the model into the application or system, making it accessible for making predictions or recommendations. Embed the model within the application or system or provide an API for seamless access.
     \item \textbf  Ensure compliance with relevant healthcare regulations and privacy standards during the model's deployment. This is crucial for safeguarding patient privacy and promoting responsible model usage. Be aware that healthcare regulations and privacy standards can vary between regions.
\end{itemize}
While deploying and integrating a fine-tuned BioBERT model into healthcare applications, consider the following:
\begin{itemize}
    \item \textbf 	Evaluate the model's performance on a held-out dataset to ensure its effectiveness with new data.
    \item \textbf 	Continuously monitor the model's performance to confirm it meets expectations.
    \item \textbf Regularly update the model to account for changes in the data.
\end{itemize}
 \subsection{Continuous Improvement:}
 Continuously update and fine-tune the model in response to new healthcare data availability or evolving clinical requirements.
 \begin{itemize}
      \item Seek feedback from healthcare professionals, leveraging their expertise in the field for model improvement. Use their insights to identify areas where the model underperforms or to uncover new potential applications. 
      \item Fine-tune the model using newly acquired healthcare data, applying the same training process employed in the model's initial training phase.
      \item Experiment with various hyperparameters to optimize the model's performance for the specific task.
      \item Apply regularization techniques to prevent overfitting, a concern that may arise when training the model on a limited dataset.
     \item Enhance the model's robustness by employing data augmentation techniques, making it more resilient to noise and data variations.
     \item Continually monitor the model's performance to ensure it meets expectations. If performance deteriorates, consider fine-tuning or updating it with fresh data.
\end{itemize}
 \subsection{Documentation and Accessibility:}
 Comprehensively document the fine-tuned BioBERT model, including pre-trained weights and code, and make it accessible to the healthcare and research community. Provide comprehensive documentation, code, and model checkpoints in various formats like a technical paper, a blog post, and a GitHub repository. This approach will expand accessibility to a broader audience.
\subsection{Ethical Considerations:}
To ensure that the fine-tuned model addresses ethical concerns related to patient privacy and data security and that it avoids inadvertently revealing sensitive patient information in compliance with healthcare regulations like HIPAA, the following specific ethical considerations should be incorporated when using a fine-tuned BioBERT model:
\begin{itemize}
  \item \textbf{Respecting Patient Privacy:}Users must refrain from utilizing the model to access or disclose sensitive patient information, including patient names, medical records, and insurance details.
 \item \textbf{Enhancing Data Security:} The model should be safeguarded against unauthorized access and use. This entails implementing measures like encryption and access control.
  \item \textbf{Mitigating Bias:} Efforts should be made to prevent bias against any particular group of people. This can be achieved by employing a balanced dataset and avoiding discriminatory features.
  \item \textbf{Ensuring Transparency:} The model must be transparent and interpretable. Users should have the capacity to comprehend how the model operates and how it generates its predictions.
 \item \textbf{Establishing Accountability:} Developers and users of the model bear responsibility for its actions. They are obligated to ensure the model's safe and responsible use.
\end{itemize}
\section{Discussion}
Based on analyzing the selected works, we realized that LLMs have the potential to revolutionize healthcare. They can introduce novel approaches to enhance clinical decision-making, facilitate information retrieval, and enable more natural language interaction. We have explored the potential benefits and limitations of integrating these language models into healthcare applications. BioBERT's primary strength resides in its capacity to comprehend and process intricate biomedical and clinical texts. Its pre-training on an extensive corpus of biomedical literature provides it with a robust foundation to accurately interpret medical terminologies, abbreviations, and concepts. Such capability proves indispensable in the healthcare context, where specialized language prevails. Moreover, BioBERT can undergo fine-tuning for specific applications, encompassing medical entity recognition, text classification, disease prediction, and question-answering. This adaptability empowers healthcare professionals to harness the model's capabilities across a broad spectrum of clinical and administrative functions.

\subsection{Advantages of using BioBERT for healthcare applications}
BioBERT offers improved clinical decision support, representing one of its most promising applications. Healthcare providers can utilize the model to swiftly access current medical knowledge, research articles, and patient records. This empowers them to render more informed decisions regarding diagnosis, treatment, and patient care, enhancing patient outcomes. BioBERT significantly enhances information retrieval efficiency from electronic health records (EHRs) and other clinical documents. Its ability to process and analyze extensive text data aids healthcare professionals in promptly accessing patient-specific information, thereby reducing the risk of overlooking critical data. The model's natural language processing capabilities make it accessible to healthcare professionals, even those without technical backgrounds. This promotes more effective communication between healthcare providers and technology, enhancing user experience and adoption. However, we must recognize and tackle the challenges linked to deploying BioBERT in healthcare:
\begin{enumerate}
  \item Data Privacy and Security: Healthcare data is highly sensitive and falls under stringent privacy regulations. To ensure BioBERT's compliance with these regulations, such as HIPAA in the United States, it is crucial to prevent data breaches and safeguard patient information.
  \item Bias and Fairness: BioBERT, like other language models, can inherit biases present in the training data. This bias can lead to disparities in healthcare if not carefully mitigated. Developing techniques to identify and rectify bias in healthcare-specific contexts is essential.
  \item Lack of Transparency: Interpreting BioBERT's decisions can be challenging due to its complex architecture. Efforts to make the model more transparent and explainable are necessary to build trust among healthcare professionals.
  \item Quality Control: Ensuring the quality and accuracy of information generated or retrieved by BioBERT is paramount. Erroneous information or recommendations could have serious consequences in clinical settings.
  \item Resource Intensiveness: Developing, fine-tuning, and maintaining BioBERT for healthcare can be resource-intensive, requiring substantial computational power, data annotation, and expert oversight.
  \item Generalization Challenges: BioBERT may struggle with generalizing to specific healthcare domains, specialties, or rare conditions if not adequately fine-tuned. Customization may be necessary to achieve optimal performance.
\end{enumerate}
BioBERT holds immense potential for revolutionizing healthcare applications by improving information retrieval, clinical decision support, and natural language interaction. However, its deployment in the healthcare sector must be accompanied by stringent measures to address privacy concerns, mitigate bias, ensure transparency, maintain data quality, allocate resources effectively, and fine-tune for specific healthcare contexts. With careful consideration and responsible implementation, BioBERT can become a valuable tool for healthcare professionals, enhancing patient care and medical research.
\section{Conclusion and Future Work}
In conclusion, this study offers valuable insights into how the Large Language Models can impact the healthcare sector. It highlights the potential of enhancing various aspects of healthcare, applications such as improving patient care and streamlining healthcare processes. However, challenges such as model performance and ethical considerations remain. Future research shall focus on addressing the existing challenges and further harnessing the capabilities of LLMs by extending to the following dimensions.
\begin{itemize}
\item Improving model performance 
\item Extending NLP for downstream tasks.
\item Harnessing the capabilities of multimodal LLMs to provide a more comprehensive understanding of patient health.
\item Developing cost-effective methods for developing and deploying LLMs.
\end{itemize}
\bibliographystyle{unsrtnat}
\bibliography{references} 
\end{document}